\def\year{2019}\relax
\documentclass[letterpaper]{article} %DO NOT CHANGE THIS
\usepackage{aaai19}  %Required
\usepackage{times}  %Required
\usepackage{helvet}  %Required
\usepackage{courier}  %Required
\usepackage{url}  %Required
\usepackage{graphicx}  %Required
\usepackage{amsfonts}
\usepackage{subfigure}
\usepackage{multirow}
\usepackage{makecell}
\usepackage{booktabs}
\begin{document}
\title{ActivityNet-QA: A Dataset for Understanding Complex Web Videos\\ via Question Answering}
\author{
Zhou Yu\textsuperscript{\rm 1},
Dejing Xu\textsuperscript{\rm 2},
Jun Yu\textsuperscript{\rm 1}\thanks{Jun Yu is the corresponding author},
Ting Yu\textsuperscript{\rm 1},
Zhou Zhao\textsuperscript{\rm 2},
Yueting Zhuang\textsuperscript{\rm 2},
Dacheng Tao\textsuperscript{\rm 3}
\\
\textsuperscript{\rm 1} Key Laboratory of Complex Systems Modeling and Simulation, \\
School of Computer Science and Technology, Hangzhou Dianzi University, Hangzhou, China.\\
\textsuperscript{\rm 2} College of Computer Science, Zhejiang University, Hangzhou, China\\
\textsuperscript{\rm 3} UBTECH Sydney AI Centre, SIT, FEIT, University of Sydney, Australia\\
\{yuz,~yujun~, yuting\}@hdu.edu.cn,~
\{xudejing,~zhaozhou,~yzhuang\}@zju.edu.cn,~
dacheng.tao@sydney.edu.au
}

\maketitle
\begin{abstract}
Recent developments in modeling language and vision have been successfully applied to image question answering. It is both crucial and natural to extend this research direction to the video domain for video question answering (VideoQA). Compared to the image domain where large scale and fully annotated benchmark datasets exists, VideoQA datasets are limited to small scale and are automatically generated, etc. These limitations restrict their applicability in practice. Here we introduce ActivityNet-QA, a fully annotated and large scale VideoQA dataset. The dataset consists of 58,000 QA pairs on 5,800 complex web videos derived from the popular ActivityNet dataset. We present a statistical analysis of our ActivityNet-QA dataset and conduct extensive experiments on it by comparing existing VideoQA baselines. Moreover, we explore various video representation strategies to improve VideoQA performance, especially for long videos. The dataset is available at \url{https://github.com/MILVLG/activitynet-qa}
\end{abstract}

\section{Introduction}
Recent developments in deep neural networks have significantly accelerated the performance of many computer vision and natural language processing tasks. These advances stimulated research into bridging the semantic connections between vision and language, such as in visual captioning \cite{donahue2015long,xu2015show}, visual grounding \cite{rohrbach2016grounding,chen2017query,yu2018rethinking} and visual question answering \cite{malinowski2015ask,fukui2016multimodal}.

Visual question answering (VQA) aims to generate natural language answers to free-form questions about a visual object (\emph{e.g.}, an image or a video). Compared to visual captioning, VQA is \emph{interactive} and provides fine-grained visual understanding. Image question answering (ImageQA) in particular has shown recent success, with many approaches  proposed to investigate the key components of this task, \emph{e.g.}, discriminative feature representation \cite{anderson2017up-down}, multi-modal fusion \cite{kim2016hadamard,yu2017mfb,yu2018beyond} and visual reasoning \cite{nam2016dual,lu2016hierarchical,johnson2017inferring}. This success has been facilitated by large scale and well annotated training datasets, such as Visual Genome \cite{krishna2016visual} and VQA \cite{antol2015vqa,goyal2017making}.

\begin{figure}
\centering
\includegraphics[width=0.47\textwidth]{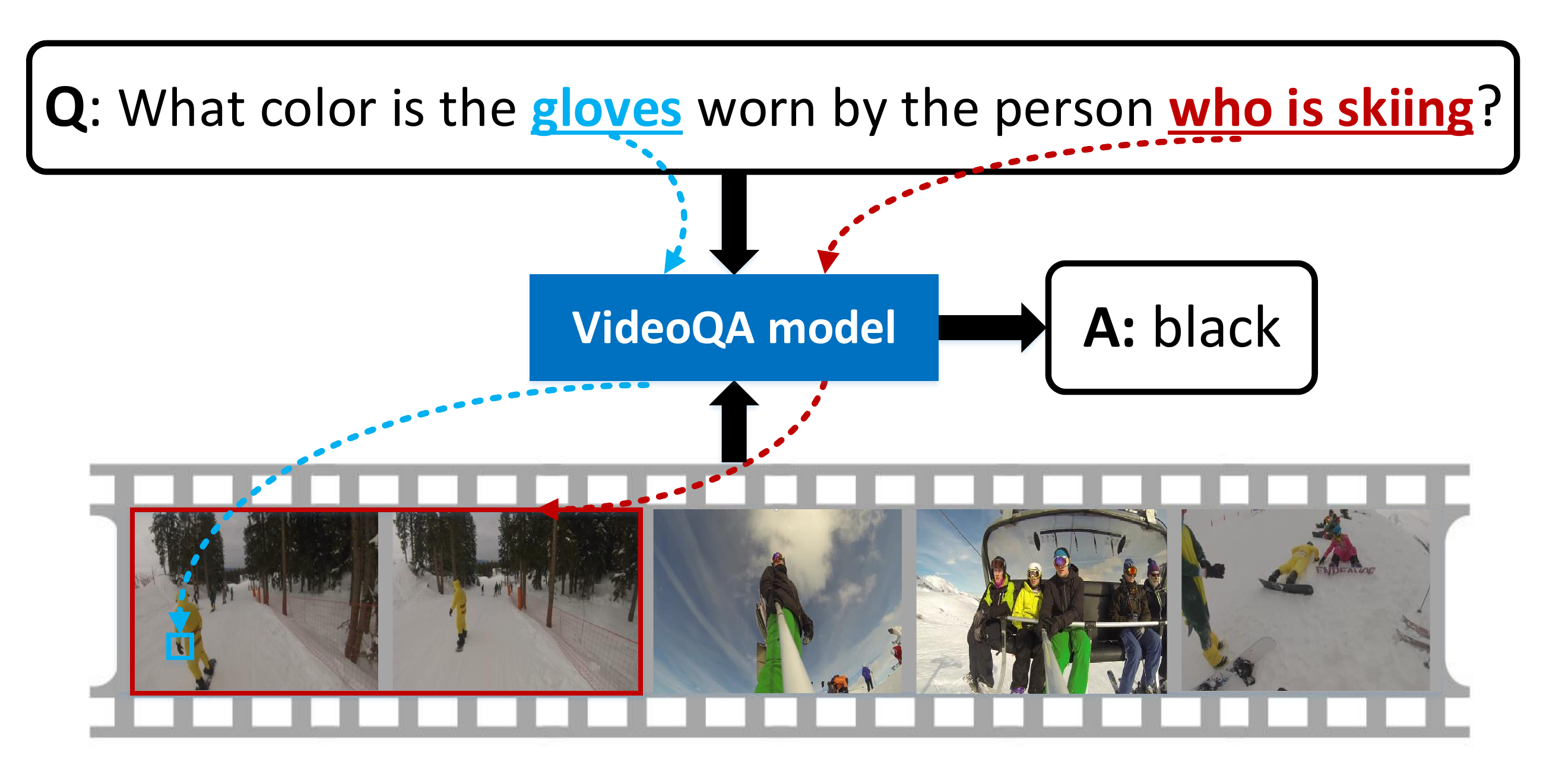}
\caption{A VideoQA example. To answer the question correctly, one should fully understand the fine-grained semantics of the questions (\emph{i.e.}, the underlined keywords) and perform spatio-temporal reasoning on the visual contents of the video (\emph{i.e.}, frames in red border and objects in blue box). }
\label{fig:example}
\end{figure}

\begin{table*}
\centering
\caption{Comparison of existing VideoQA datasets with ours (OE: open-ended, and MC: multiple-choice).}
\small
\label{table:dataset_compare}
\begin{tabular}{c|cccccc}
\toprule
Datasets & \makecell{Video source} & \makecell{QA pairs \\generation}& QA tasks & $\#$ \makecell{Videos} & $\#$ QA pairs & \makecell{Average\\ video length} \\
\midrule
MSVD-QA \cite{xu2017video} & MSVD & Automatic &OE &1,970 & 50,505 & 10s\\
MSRVTT-QA \cite{xu2017video}& MSRVTT & Automatic&OE& 10,000 & 243,680 & 15s\\
TGIF-QA \cite{jang2017tgif}& TGIF & Automatic $\&$ Human & OE $\&$ MC & 56,720 & 103,919 & 3s \\
MovieQA \cite{tapaswi2016movieqa} & Movies & Human & MC &6,771& 6,462 & 200s  \\
Video-QA \cite{zeng2017leveraging}  &Jukinmedia & Automatic&OE&18,100&174,775& 45s\\
\midrule
ActivityNet-QA (Ours) & ActivityNet  & Human & OE & 5,800& 58,000& 180s \\
\bottomrule
\end{tabular}
\end{table*}

Video question answering (VideoQA) can be seen as a natural but more challenging extension of ImageQA, due to the additional complexity of understanding of image sequences and more diverse types of questions asked. Figure \ref{fig:example} shows an example of VideoQA. To accurately answer the question, a VideoQA model requires simultaneous fine-grained video content understanding and spatio-temporal reasoning. Existing approaches mainly focus on the temporal attention mechanism \cite{jang2017tgif,xu2017video} or memory mechanism \cite{na2017read,kim2017deepstory,zhao2018multi}. Na \emph{et al.} introduced a read-write memory network to fuse multi-modal features and store temporal information using a multi-stage convolutional neural networks model \cite{na2017read}. Xu \emph{et al.} represented a video as appearance and motion stream features and introduced a gradually refined attention model to fuse the two-stream features together. \cite{xu2017video}. Gao \emph{et al.} proposed a co-memory network to jointly model and interact with the motion and appearance information \cite{gao2018motion}. Zhao  \emph{et al.} introduced an adaptive hierarchical encoder to learn the segment-level video representation with adaptive video segmentation, and devised a reinforced decoder to generate the answer for long videos \cite{zhao2018open}.

As noted above, high-quality datasets are of considerable value for VQA research. Several VideoQA datasets have been compiled for different scenarios, such as MovieQA \cite{tapaswi2016movieqa}, TGIF-QA \cite{jang2017tgif}, MSVD-QA, MSRVTT-QA \cite{xu2017video}, and Video-QA \cite{zeng2017leveraging}. Most of these VideoQA datasets exploit video source data from other datasets and then add question-answer pairs to them. The detailed statistics of these datasets are listed in Table \ref{table:dataset_compare}. We can see that these existing datasets are imperfect and have at least one of the following limitations:

 \begin{itemize}
   \item The datasets are small scale. Without sufficient training samples, the obtained model suffers from under-fitting. Without sufficient testing samples, the evaluated results are unreliable.
   \item The questions and answers are automatically generated by algorithms (\emph{e.g.}, obtained from the captioning results or narrative descriptions using off-the-shelf algorithms) rather than human annotation. Automatically generated question-answer pairs lack diversity, making the learned model easy to over-fit.
   \item The videos are short. The length of a video is closely related to the complexity of video content. Questions on short videos (\emph{e.g.}, less than 10 seconds) are usually too easy to answer making it difficult distinguish the performance of different VideoQA approaches on the dataset.
   \item The videos represent a small number of activities. This severely restricts the generalizability of the VideoQA models trained on these datasets and poorly reflects model performance in real-world use.
 \end{itemize}

In this paper, we construct a new benchmark dataset \emph{ActivityNet-QA} for evaluating VideoQA performance. Our dataset exploits 5,800 videos from the ActivityNet dataset, which contains about 20,000 untrimmed web videos representing 200 action classes \cite{caba2015activitynet}. We annotate each video with ten question-answer pairs using crowdsourcing to finally obtain 58,000 question-answer pairs. Compared with other VideoQA datasets, ActivityNet-QA is of large scale, fully annotated by humans, and with very long videos. To better understand the properties of ActivityNet-QA, we present statistical and visualization analyses. We further conduct experiments on ActivityNet-QA and compare results produced by existing VideoQA baselines.

\section{ActivityNet-QA Dataset}
We first introduce the ActivityNet-QA dataset from three perspectives, namely \emph{video collection}, \emph{QA generation}, and \emph{statistical analysis}.

\subsection{Video Collection}
We first collect videos for the dataset. Due to time limitation, we are unable to annotate every video in ActivityNet. Instead, we sample 5,800 videos from the 20,000 videos in ActivityNet dataset. Specifically, we sample 3,200/1,800/800 videos from the original {train}/{val}/{test} splits of ActivityNet respectively. Moreover, we take class diversity and balance into consideration. Since the videos in the {train} and {val} splits of ActivityNet possess class labels, we use this information as a prior to guide sampling and force a uniform distribution of class labels in the sampled videos. The class information is not available for the test split, so we adopt a simple random sampling strategy instead.

\subsection{QA Generation}

\begin{figure*}
\centering
\subfigure[QA length distributions] {\includegraphics[width=0.24\linewidth]{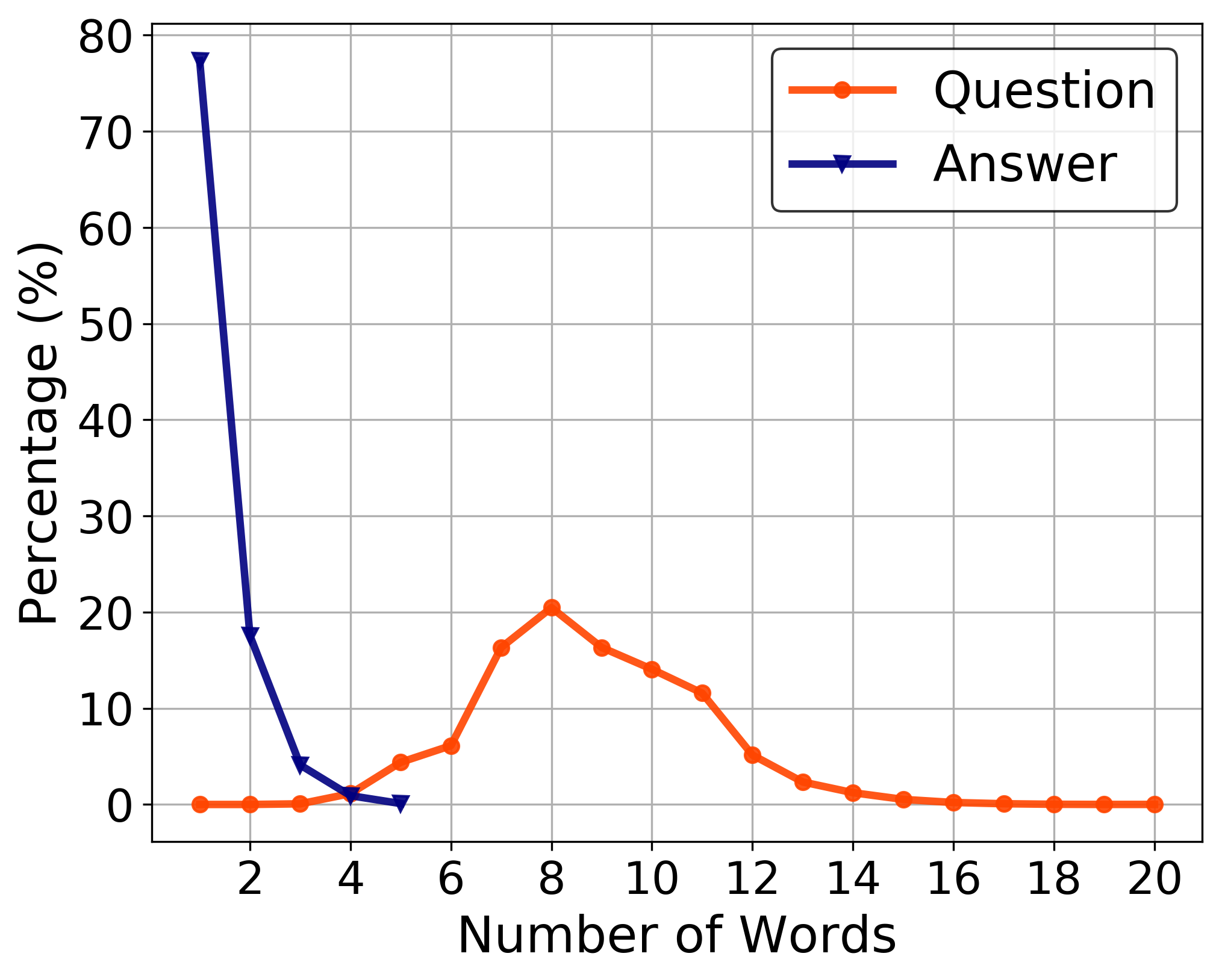}\label{fig:qa_length_dist}}
\subfigure[Average QA lengths] {\includegraphics[width=0.23\linewidth]{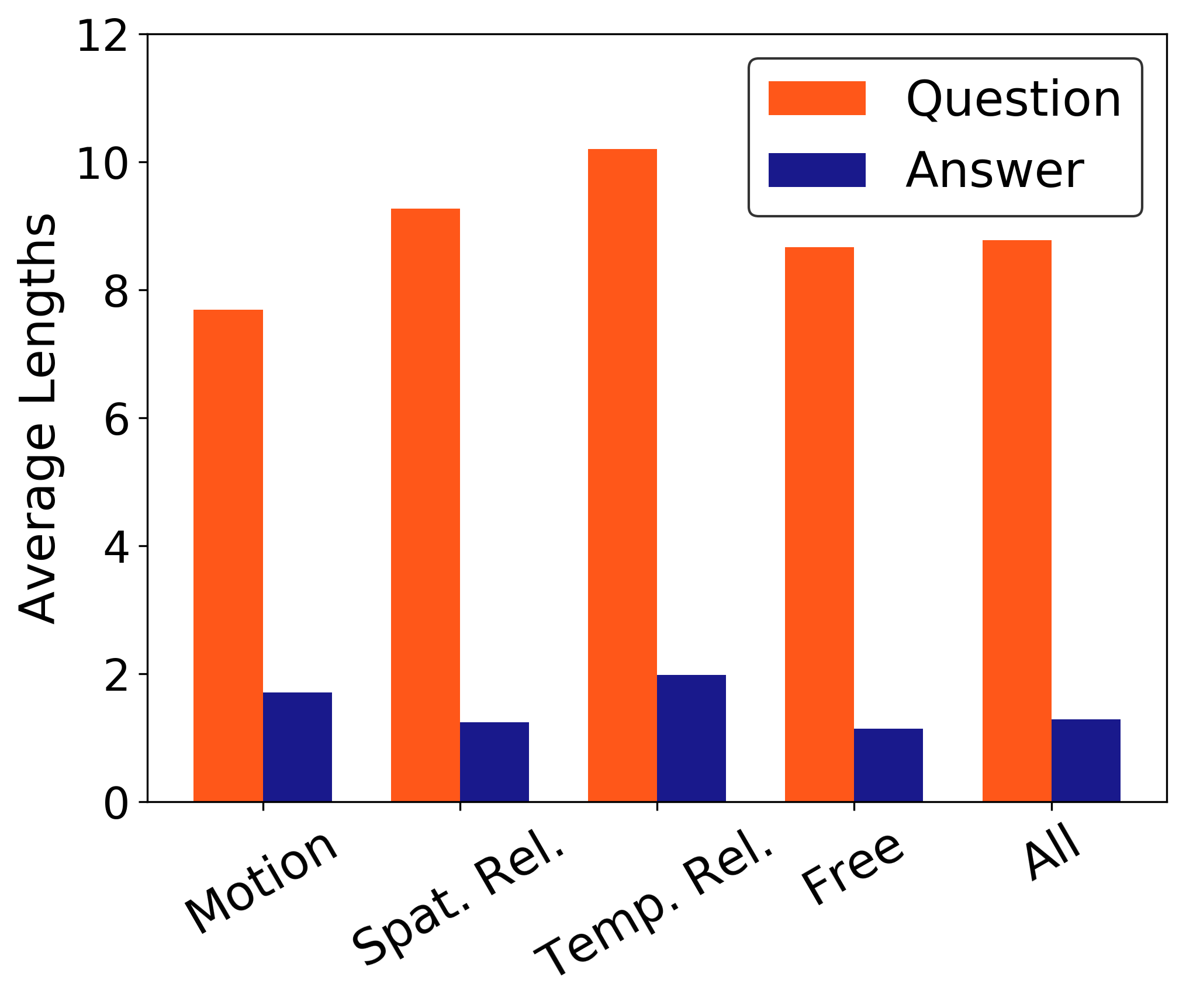}\label{fig:qa_length_type}}
\subfigure[Question distribution] {\includegraphics[width=0.22\linewidth]{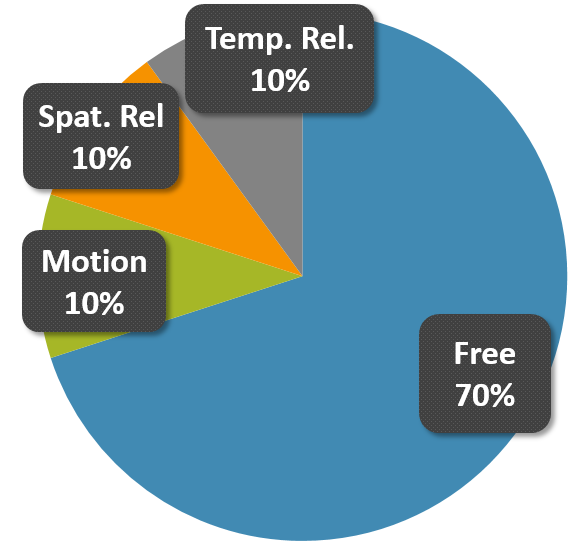}\label{fig:ques_dist_all}}
\subfigure[Answer distribution for \emph{Free} type] {\includegraphics[width=0.265\linewidth]{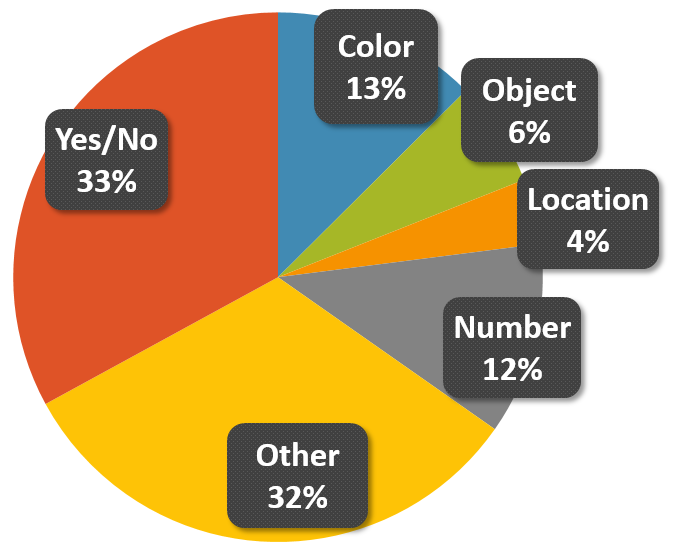}\label{fig:free_dist_all}}
\caption{The statistics of our ActivityNet-QA dataset. }
\label{fig:div_comparasion}
\end{figure*}
As the videos are collected, we generate QA pairs for each video. To reduce the labor costs, some VideoQA datasets exploit the narrative descriptions or captions of videos \cite{jang2017tgif,xu2017video}, to automatically generate QA pairs using off-the-shelf algorithms \cite{ren2015exploring}. However, since the textual descriptions contains relatively little information about the videos, these generated QA pairs lack diversity and are often redundant. Therefore, we generate QA pairs by human crowdsourcing.

To control the generated questions, we define three template question types and ask the annotators to cover all the three question types for every video. Beyond these three questions, annotators are free to ask arbitrary questions about the videos. The three question types are as follows:
\\
\textbf{Motion.} This type of question interrogates coarse temporal action understanding. Compared with traditional action recognition, this task is more challenging in this setting. To correctly answer the question with respect to a long video, a VideoQA model needs to correctly localize the action referred to by the question.
\\
\textbf{Spatial Relationship.} This type of question tests spatial reasoning on one static frame. In contrast to the spatial reasoning in ImageQA \cite{johnson2017clevr}, this task additionally examine the temporal attention ability to find to the frame from the whole video first.
\\
\textbf{Temporal Relationship.} As a companion to spatial relationship, this type of questions examines the ability of reasoning temporal relationships of objects from a sequence of frames. As a prerequisite, one should find the related frames from the whole video first.

For the free type questions, it is hard to classify them into non-overlapped types even for humans. Referring to the taxonomy in existing VQA datasets \cite{ren2015exploring,antol2015vqa}, we manually categorize the samples into the following six classes by their answer types: \emph{Yes/No}, \emph{Number}, \emph{Color}, \emph{Object}, \emph{Location} and \emph{Other}.

To control the quality of generated questions, the following practical principles are applied:
\begin{itemize}
  \item Questions and answers that are too long are probably caused by improper representation. Therefore, we empirically restricted the maximum question length to 20 words and maximum answer length to 5 words.
  \item For each QA pair, the question annotator and answer annotator are separate. If the answer annotator regards the generated question \emph{unanswerable}, this question is double-checked and may have been further regenerated. Employing this strategy effectively improve question objectivity, which is important for obtaining high-quality annotations.
  \item A portion of questions are randomly selected and sent to multiple annotators. The multiple answers to one question are merged by majority voting. Employing this strategy reduced the probability of erroneous answers and evaluated annotator reliability.
\end{itemize}

\begin{table}
\centering
\caption{Examples of questions in different types. }
\scriptsize
\label{table:question_examples}
\begin{tabular}{c|l}
\toprule
Types & Questions \\
\midrule
\multirow{3}{*}{\makecell{Motion}} & What are the person wearing earphones doing?  \\
& What are people doing at the beginning of the video?\\
& What is person wearing red t-shirt doing?\\
\midrule
\multirow{3}{*}{\makecell{Spat. Rel.}} & What is on the left of the lawn? \\
& What is on the left side of the man on his knees?\\
& What is behind of the person sitting in the video?\\
\midrule
\multirow{3}{*}{\makecell{Temp. Rel.}} & What happened to the person in black before falling down?\\
& What happened to the woman before drying her hair?  \\
& What happened to the person before playing violin?\\
\midrule
\multirow{5}{*}{\makecell{Free}} & How many people are there in the video? [Number] \\
& Is the athlete in the room? [Yes/No]\\
& What are the animals that appear in the video? [Object]\\
& What is the color of the person's pants? [Color]\\
& Where is the person in a black coat? [Location]\\
& What is the gender of the athlete? [Other]\\
\bottomrule
\end{tabular}
\end{table}

The initial QA pairs are in Chinese, since our crowdsourcing platform is located in China. As the lengths of questions and answers are well controlled, the state-of-the-art machine translation algorithms can easily translate them into English. To further improve the quality of the translated results, we use a novel strategy to automatically detect potential mistakes in the translated results. For each sentence in Chinese, we use the APIs from the four commercial translation engines of Google, Baidu, Sogou and Youdao to obtain four translated versions in English. We evaluate the average similarities of each two natural language sentences using CIDEr score \cite{vedantam2015cider}, setting an empirical threshold to the average CIDEr score, and manually checking samples that did not reach the threshold. For the remaining samples that did reach the threshold, we use the results obtained from the Baidu translation engine.

To better understand the different question types and the quality of translations, some examples from our dataset are shown in Table \ref{table:question_examples}.

\begin{figure*}
\centering
\includegraphics[width=0.97\textwidth]{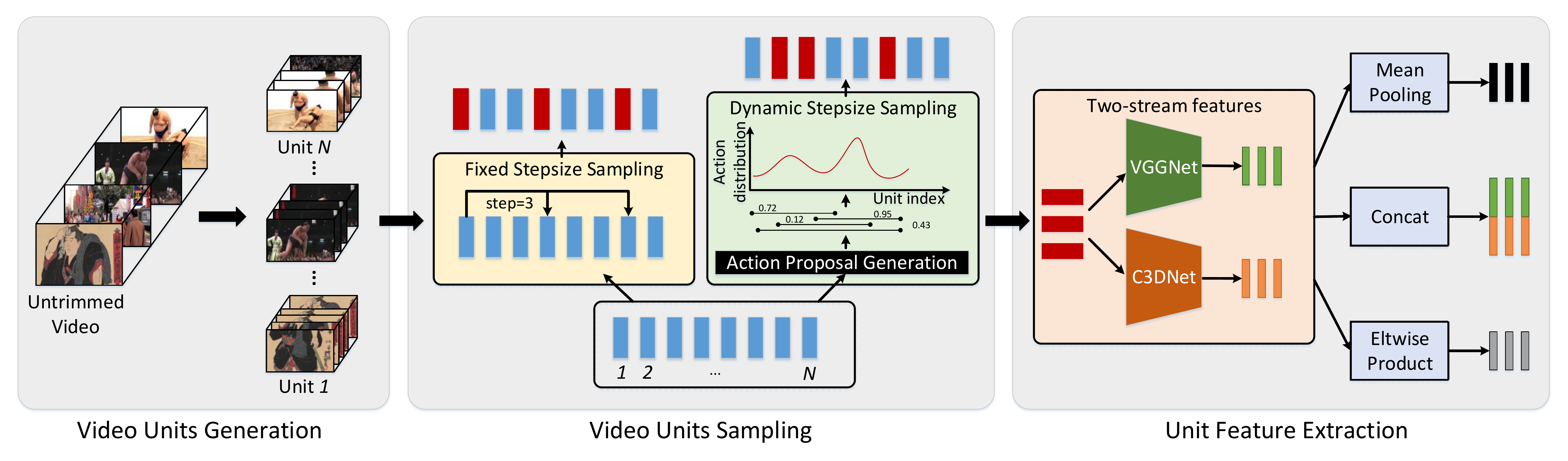}
\caption{The flowchart of video feature representation procedures, including video units generation (left), video units sampling (middle) and unit feature extraction (right). }
\label{fig:flowchart}
\end{figure*}

\subsection{Statistical Analysis}
Here we present the detailed statistics of our ActivityNet-QA dataset. The distributions of question and answer lengths are shown in Figure \ref{fig:qa_length_dist}. As noted above, the maximum question length is 20 and the maximum answer length is 5 respectively (in English). The average QA lengths for all the question types are reported in Figure \ref{fig:qa_length_type}. Regardless of question type, the average question length is 8.67 and average answer length is 1.85. Similar to \cite{antol2015vqa}, the answer lengths in our dataset are relatively short. Short answers are easier to process, and one can simply treat the answering problem as multi-class classification.

We also investigate the distribution of the questions (Figure \ref{fig:ques_dist_all}). For each video, we generate exactly ten QA pairs, including one \emph{motion} type question, one \emph{spatial relationship} type question, and one \emph{temporal relationship} type question, respectively. The remaining seven questions are classified as \emph{free} type as they are generated without constraints.
To eliminate the effect of the answer prior and improve the role of video understanding for \emph{Yes/No} type questions, we balance the ratio of \emph{yes} and \emph{no} samples to make it close to one.
To better understand the organization of the \emph{free} type questions, the answer distribution is shown in Figure \ref{fig:free_dist_all}.

\section{Methods}
In this section, we explore the difficulty of the ActivityNet-QA dataset using several baseline models based on different types of video features.

\subsection{Video Feature Representation}

Existing VideoQA approaches usually extract \emph{two-stream} features from the motion channel and the appearance channel of video, respectively \cite{xu2017video}.

Since the untrimmed videos are very long, we split the videos into small units, each unit containing 16 consecutive frames without overlap between any two units. The average number of units counted on all the videos is 270, which is still too large for existing VideoQA models. Besides, the number of units varies for different videos, further complicating model training. Therefore, we propose two alternative sampling strategies to sample a fixed number of units $T$ for all videos:
\\
\textbf{Fixed Stepsize (FS).} For a video, assume it contains $N$ units and we expect to output $T$ units. This strategy evenly sample $T$ units from the $N$ units with a fixed stepsize $\lceil\frac{N}{T}\rceil$.
\\
\textbf{Dynamic Stepsize (DS).} An untrimmed video may contain many worthless frames with little information. To make the sampled video units more discriminative, we propose a sampling strategy with dynamic stepsize such that the selected units have a high probability of containing meaningful actions. To achieve this, we first introduce an external temporal action proposal model \cite{caba2016fast} to generate a set of action proposals:
\begin{equation}
P=\{(t^{\mathrm{start}}_i, t^{\mathrm{end}}_i, c_i)\}
\end{equation}
where each proposal $p_i\in P$ contains a start index $t^{\mathrm{start}}_i\in\mathbb{R}$, an end index $t^{\mathrm{end}}_i\in\mathbb{R}$ and a confidence score $c_i\in\mathbb{R}$. For each video unit, we regard it as a bin in a histogram. If the duration of a proposal $p_i$ covers the video unit, its confidence score $c_i$ is added to this unit. After traversing the candidate set $P$, we obtain a histogram w.r.t the video units. By normalizing the histogram using the softmax function, we obtain an action score distribution over the video units indicating the probability that one unit has valid actions. Finally, we sample the units w.r.t the score distribution to obtain $T$ units with dynamic stepsize.

As we have obtained the sampled video unit set, for each unit $u_i$, we used the VGG-16 network pre-trained on the ImageNet dataset \cite{simonyan2014very} to extract the appearance feature (the \texttt{fc7} feature $x_i\in\mathbb{R}^{4096}$) given the central frame of the unit, and the C3D network pre-trained on the Sport-1M dataset \cite{tran2015learning} to extract the motion features (the \texttt{fc7} feature $y_i\in\mathbb{R}^{4096}$) given the whole 16 consecutive frames of the unit.

To fuse the two-stream features, we use three fusion strategies: \emph{Mean Pooling}, \emph{Concat} and \emph{Eltwise Product} to obtain the 4096-D, 8192-D, and 4096-D fused visual features for each video unit, respectively. We then perform $L_2$ normalization on each fused visual feature.

The overall flowchart for video feature representation is illustrated in Figure \ref{fig:flowchart}.

\begin{table*}
\centering
\caption{The accuracies of the methods in different question types. Q-type prior denotes a simple baseline using the most popular answer per question type as the prediction.}
\label{table:compare_baselines}
\begin{tabular}{c|ccccccccc}
\toprule
\multirow{3}{*}{\makecell{Methods}} & \multicolumn{6}{c}{Accuracy ($\%$)} && \multicolumn{2}{c}{WUPS ($\%$) }\\
\cmidrule{2-7}
\cmidrule{9-10}
 &Motion & \makecell{Spat. Rel.} & \makecell{Temp. Rel.} & Free && All && WUPS@0.9 & WUPS@0.0\\
\cmidrule{1-5}
\cmidrule{7-7}
\cmidrule{9-10}
\makecell{Q-type prior} & 2.9 & 5.8 & 1.4 &    19.7 &    & 14.8   &&   16.4   &     35.1 \\
\midrule
E-VQA & 2.5&6.6&1.4&34.4&&25.1&&29.3&53.5 \\
E-MN  &3.0&8.1&1.6&36.9&&27.1&&31.5&55.9 \\
E-SA  & \textbf{12.5} & \textbf{14.4} & \textbf{2.5} & \textbf{41.2} && \textbf{31.8} && \textbf{34.9} & \textbf{56.4}\\
\bottomrule
\end{tabular}
\end{table*}

\subsection{VideoQA Baselines}
Based on the extracted visual features, we implement the following VideoQA baselines. Note that the focus of this paper is the constructed ActivityNet-QA dataset and a discussion of what influences the performance on the dataset. Therefore, we do not perform comparison with complex VideoQA models, such as \cite{gao2018motion,xu2017video}.

\textbf{E-VQA} is the extension of an ImageQA baseline \cite{antol2015vqa}, where one long-short term memory (LSTM) network \cite{hochreiter1997long} is used to encode all words in the question and another different LSTM network is used to encode the frames in the video. The features of the question and videos are then fused into the joint feature representation with element-wise multiplication for answer prediction.

\textbf{E-MN} is the extension of the end-to-end memory networks model \cite{sukhbaatar2015end} for ImageQA, where the bidirectional LSTM networks are used to update the frame representations of the video. The updated representations are mapped into the memory and the question representation is used to perform multiple inference steps to predict the answer.

\textbf{E-SA} is the extension of the soft attention model \cite{yao2015describing} for ImageQA, where the question are fist encoded using a LSTM network. The encoded question feature is used to attend on features of video features. Finally, the question feature and weighted video feature are fused to predict the answer.

All the above baselines are trained in an end-to-end manner. Since they are decoupled from the video features, they can be flexibly combined with the video features obtained by different strategies.

\section{Experiments}
We evaluate the aforementioned VideoQA models on our ActivityNet-QA dataset. We use 3,200 videos and 32,000 corresponding QA pairs in the {train} split to train the models, and 1,800 videos and 18,000 corresponding QA pairs in the {val} split to tune hyper-parameters. We report the predicted results on 800 videos and 8,000 QA pairs in the {test} split.

\subsection{Experimental Setup}
We formulate the VideoQA problem as a multi-class classification problem with each class corresponding to an answer. To generate the answer vocabulary, we choose the top 1,000 most frequent answers in the train split as the answer vocabulary, which covers 84.7$\%$ / 86.2$\%$ / 85.6$\%$ of the train/val/test answers, respectively. To generate the question vocabulary, we select the top 8,000 most frequent words from the questions in the {train} split. We take the token \emph{unk} for out-of-vocabulary words. Since there are several video feature representation strategies, we use the FS sampling with $T=20$ and the mean-pooling fusion strategies as the default options in the experiments unless otherwise stated.

We implement all the methods and train the models using TensorFlow. For all models, we use the Adam solver with a base learning rate $\alpha=0.001$, $\beta_1=0.9$, and $\beta_2=0.99$ and train the models to up to 100 epochs with a batch size of 100. The early stopping strategy is used if the accuracy on the validation set does not improve for 10 epochs. All models use the pre-trained 300-dimensional GloVe embedding \cite{pennington2014glove} to initialize the question embedding layer. For the models using LSTM networks, the number of LSTM hidden units is set to 300, and the common space dimension is set to 256 as suggested by \cite{xu2017video}. The number of memory units for E-MN is set to 500 as suggested by \cite{zeng2017leveraging}.

\subsection{Evaluation Criteria}
We evaluate the performance using two common evaluation criteria for VideoQA, \emph{i.e.}, accuracy \cite{xu2017video} and WUPS \cite{malinowski2014multi}. For the QA pairs in the test set with size $N$, given any testing question $\mathbf{q}_i\in Q$ and its corresponding ground-truth answer $\mathbf{y}_i\in Y$, we denote the predicted answer from the VideoQA model by $\mathbf{a}_i$. Note that $\mathbf{a}_i$ or $\mathbf{y}_i$ corresponds to a sentence which can be seen as a set of words. Based on the definition above, the two evaluation criteria are:

\textbf{Accuracy} is a criterion that is used to commonly used to measure the performance of classification tasks.
\begin{equation}
\textrm{Accuracy} = \frac{1}{N}\sum\limits_{i=1}^N \mathbf{1}[\mathbf{a_i}= \mathbf{y_i}]
\end{equation}
where $\mathbf{1}[\cdot]$ is an indicator function that accuracy of the sample is 1 only if $\mathbf{a_i}$ and $\mathbf{y_i}$ are identical, and 0 otherwise.

\textbf{WUPS} is a generalization of the accuracy measure that accounts for word-level ambiguities in the answer words. The WUPS score with the threshold $\gamma$ is given by
\begin{small}
\begin{equation}
\begin{array}{rcl}
\textrm{WUPS} = \frac{1}{N}\sum\limits_{i=1}^N\textrm{min}\{\prod\limits_{w\in \mathbf{a}_i}\mathop{\mathrm{max}}\limits_{v\in \mathbf{y}_i}\mu_{\gamma}(w,v),
\prod\limits_{v\in \mathbf{y}_i}\mathop{\mathrm{max}}\limits_{w\in \mathbf{a}_i}\mu_{\gamma}(w,v)\}
\end{array}
\end{equation}
\end{small}
where $w$ and $v$ are the words in the each predicted answer and ground-truth answer respectively. $\mu_{\gamma}(w,v)$ is given by
\begin{equation}
\mu_{\gamma}(w,v)=\left\{
\begin{array}{rcl}
\mathrm{WUP}(w,v) & \textrm{if~}\mathrm{WUP}(w,v) \geq \gamma \\
0.1\cdot \mathrm{WUP}(w,v) & \textrm{otherwise}\\
\end{array}
\right.
\end{equation}
Following the setting in \cite{malinowski2015ask}, we choose two thresholds $\gamma=0.0$ and $\gamma=0.9$ for calculating the WUPS score and denote them by WUPS@0.0 and WUPS@0.9, respectively.

\begin{figure*}
\centering
\subfigure[Sampling strategies] {\includegraphics[width=0.28\linewidth]{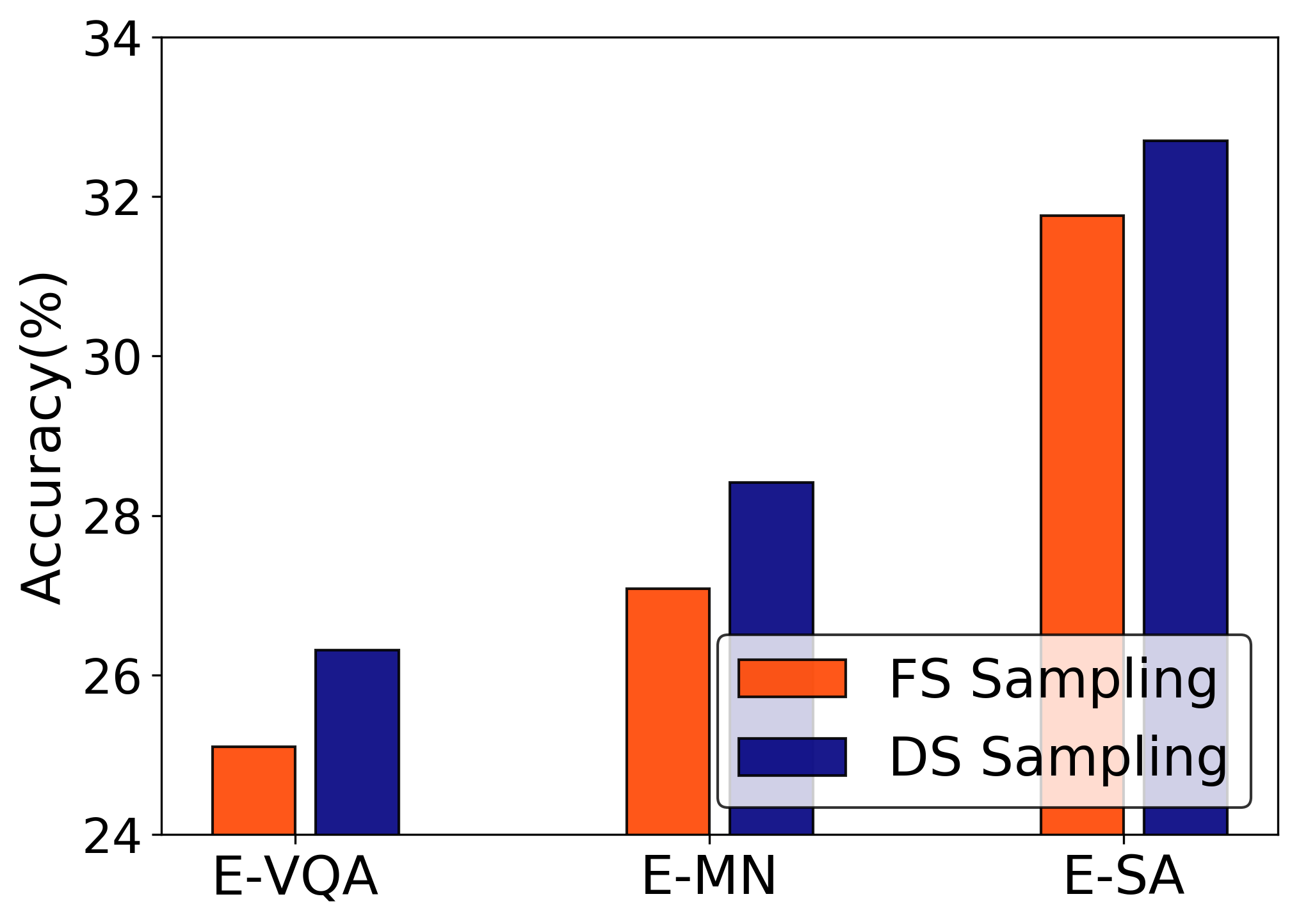}\label{fig:sampling}}
\subfigure[Sampling frequencies] {\includegraphics[width=0.28\linewidth]{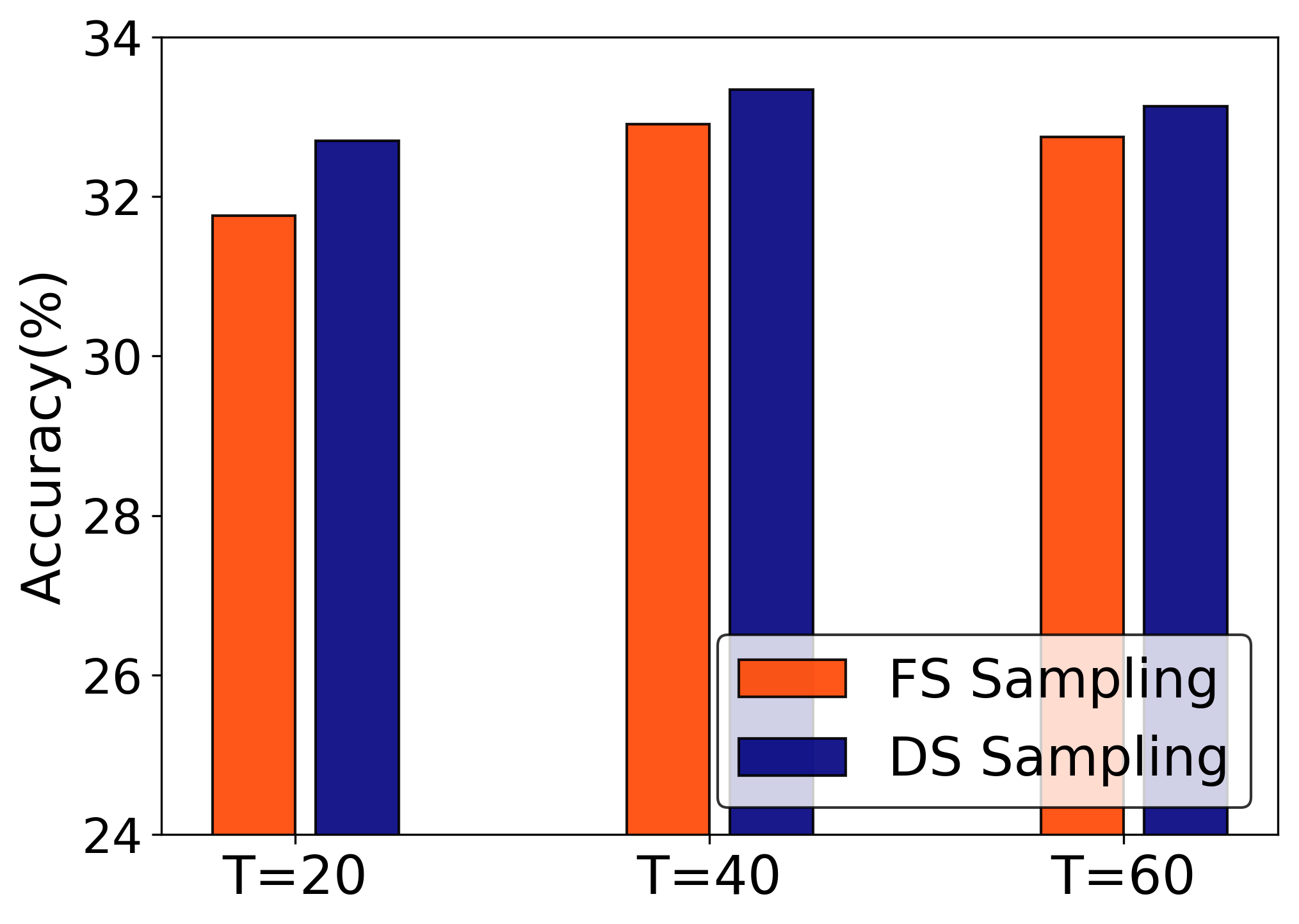}\label{fig:vary_T}}
\subfigure[Fusion strategies] {\includegraphics[width=0.28\linewidth]{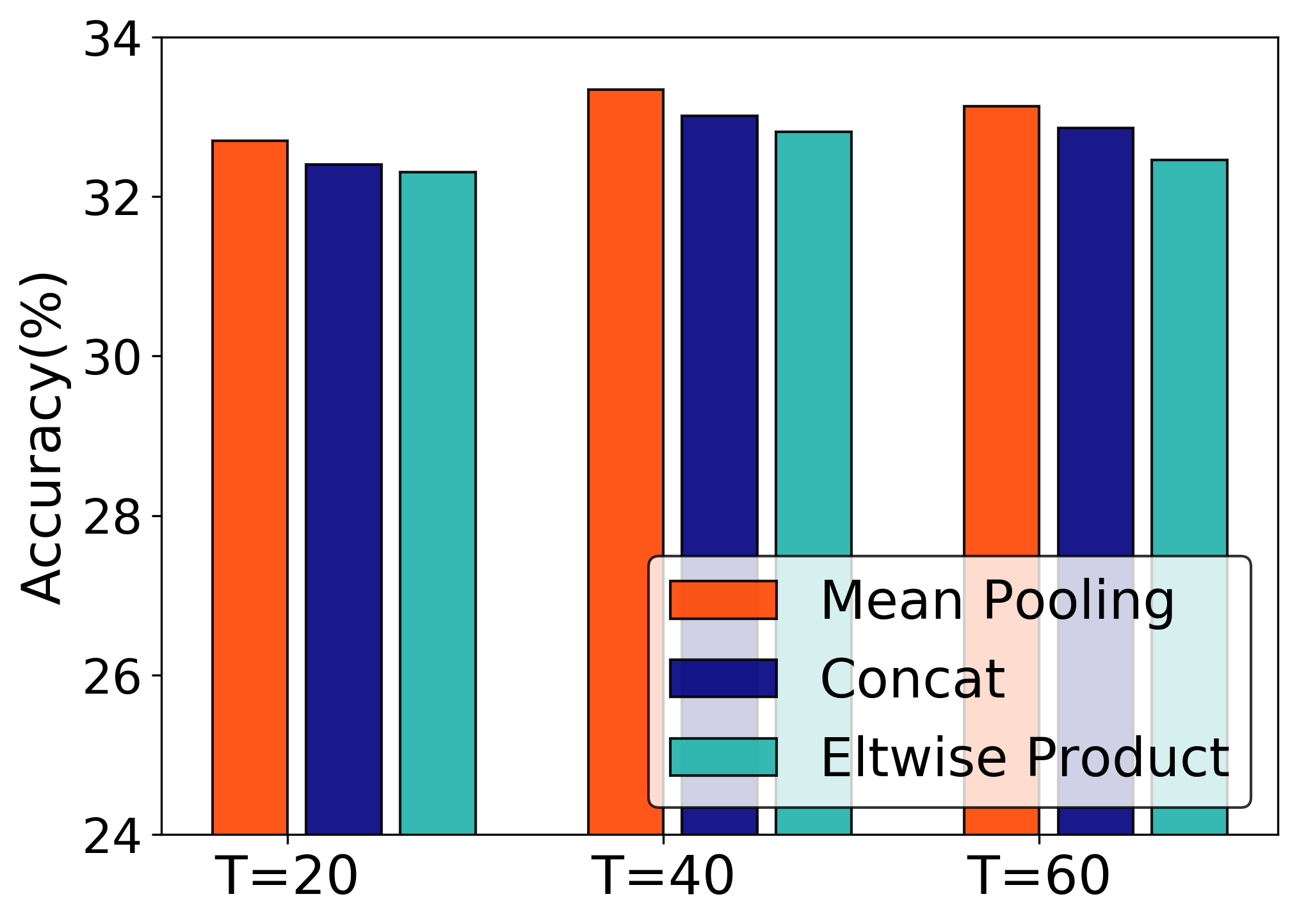}\label{fig:fusion}}
\caption{Overall accuracies of different strategies in video feature representations. (a) sampling strategies w.r.t. different VideoQA methods; (b) sampling frequencies w.r.t. different sampling strategies for E-SA; (c) sampling frequencies w.r.t. different fusion strategies for E-SA.}
\label{fig:ablation}
\vspace{-10pt}
\end{figure*}

\begin{figure*}
\centering
\includegraphics[width=0.9\textwidth]{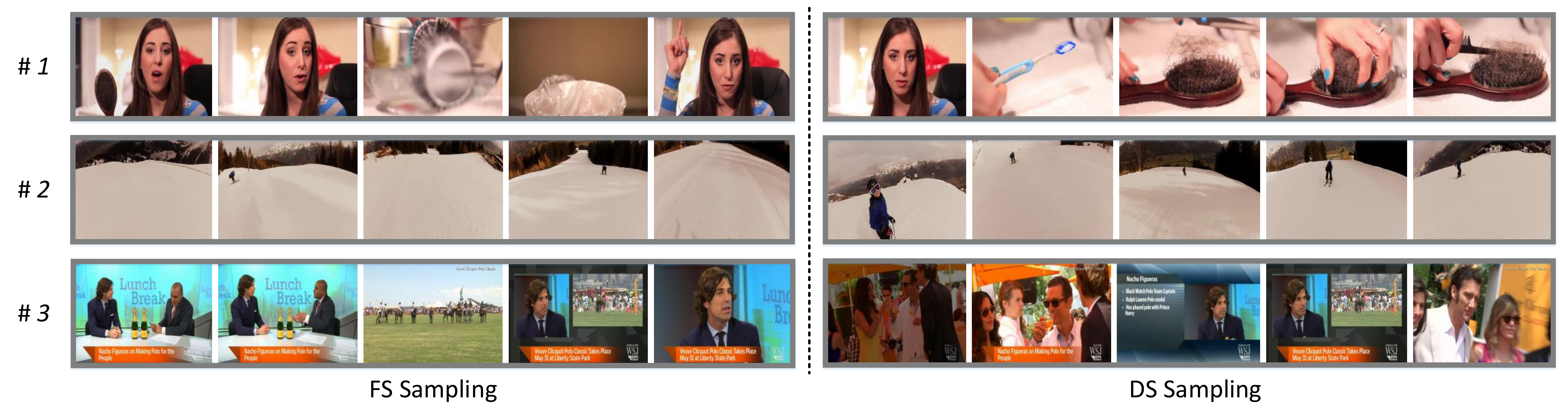}
\caption{Visualizations of three video examples with two sampling strategies (DS sampling on the left and FS sampling on the right). Each row shows the sampled video units (represented by their central frames) for a video with sampling frequency $T$=5.}
\label{fig:vis_sampling}
\vspace{-10pt}
\end{figure*}

\subsection{Results and Discussion}

Table \ref{table:compare_baselines} shows the performance of the baselines on all question types based on the two evaluation criteria. From these results, we can make the following observations: 1) all baselines significantly outperform the \emph{Q-type prior} baseline with respect to the overall accuracy and WUPS scores, indicating that without understanding the visual content of videos, one cannot achieve good performance on our dataset; 2) the accuracy of the \emph{temporal relationship} type is lower than the others. This can be explained by the fact that temporal reasoning over long videos is still not well solved by baselines, and there remains significant room for further improvement; 3) E-SA slightly outperforms E-VQA and E-MN both in terms of accuracy and WUPS, respectively. However, the overall performance is still far from satisfactory, reflecting the difficulty of the dataset.

Table \ref{table:free_type} provides the detailed accuracies for the free type questions. The accuracy for the \emph{Object} class is lower than the others due to the diversity of possible answers. E-SA still outperforms the other two methods steadily, indicating its effectiveness in modeling temporal attention.

\begin{table}
\centering
\caption{The detailed accuracies of the Free type questions.}
\label{table:free_type}
\begin{tabular}{c|cccccc}
\toprule
 & Y/N & Color & Obj.& Loc.  & Num. & Other \\
\midrule
E-VQA &52.7 &	27.3&	7.9&	8.8&	44.2&	20.6	\\
E-MN &55.1&	28.0&	12.0&	12.2&	44.4&	24.2\\
E-SA & \textbf{59.4} & \textbf{29.8} & \textbf{14.2} & \textbf{25.9} & \textbf{44.6} & \textbf{28.4}\\
\bottomrule
\end{tabular}
\end{table}

\begin{figure*}
\centering
\includegraphics[width=0.94\textwidth]{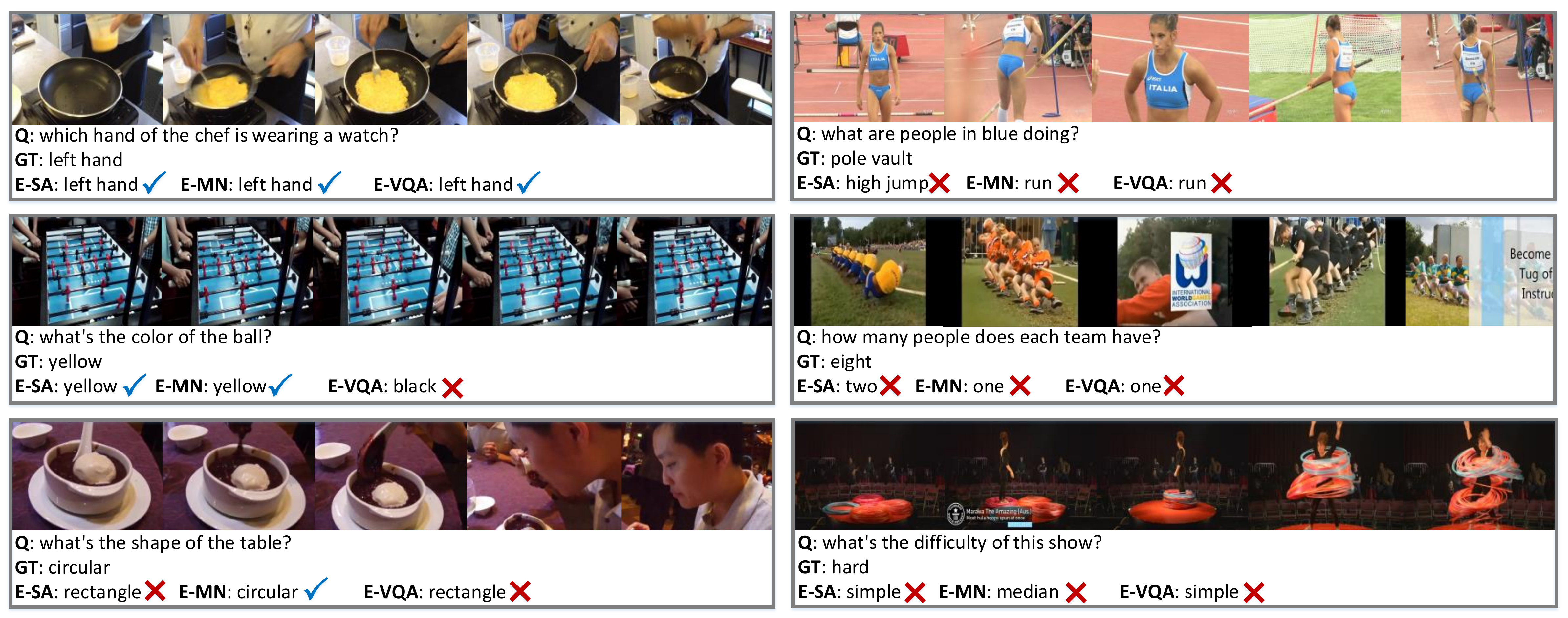}
\caption{Visualizations of the results obtained by different methods. For each video example, we show the questions (Q), ground-truth answers (GT) and the predictions of different methods, respectively. The left column shows the examples that at least one method give correct predictions, while the right column shows the examples that all methods give wrong predictions.}
\label{fig:vis_results}
\end{figure*}
We next investigate the effect of using different video feature representation strategies.
\\
\textbf{Sampling strategies.} The effect of using different sampling strategies for different methods is shown in Figure \ref{fig:sampling}. The results show that the performance of all baselines improved by at least 1$\%$ when the FS sampling is replaced with the DS sampling. This verifies that the action distribution is an important prior when extracting video features, especially when the videos are long. The sampled frames by the DS sampling can be seen as the \emph{key-frames}, which better reflect fine-grained video semantics. To better understand the differences between the two sampling strategies, we visualize the video units (represented by their central frames) in Figure \ref{fig:vis_sampling}. It can be seen that the DS sampling obtains more representative and diverse video units compared to the FS sampling.
\\
\textbf{Sampling frequencies.} Figure \ref{fig:vary_T} shows the effect of $T$=$\{20, 40, 60\}$ for E-SA. As the sampling strategy is correlated with the sampling frequency, we report the accuracies with respect to different sampling strategies. The results show that as $T$ increases, the performance gap between the fixed sampling and dynamic sampling narrows. This can be interpreted as denser sampling better preserve the detailed information in videos. Moreover, using the video features generated with dense sampling frequency (\emph{e.g.}, $T$=60) greatly increase the complexity of VideoQA models, leading to degraded performance.
\\
\textbf{Fusion strategies.} In Figure \ref{fig:fusion}, we explore the effect of using different fusion strategies with ($T$=$\{20,40,60\}$) and FS sampling for E-SA. It can be seen that \emph{Mean Pooling} achieves the best performance compared to other two fusion strategies in terms of accuracy and robustness.

For qualitative analysis, we present some successful and failed cases obtained by different methods in Figure \ref{fig:vis_results}. These methods show a greater probability of correctly answering questions that focused on static frame, but fail to answer the questions involving temporal reasoning. These observations are useful for guiding further improvements for VideoQA models in the future.

\section{Conclusion and Future Work}
In this paper, we present a new large scale dataset \emph{ActivityNet-QA} for understanding complex web videos by question answering. Compared with existing VideoQA datasets, our dataset is unique in that: 1) the videos originate from ActivityNet, a large-scale video understanding dataset with long web videos; 2) the QA pairs are fully annotated by crowdsourcing. To guarantee the quality of our dataset, we conduct significant pre- and post-processing by both algorithmic and human efforts. Based on the constructed dataset, we apply several baselines to analyze the difficulty of our dataset and also investigate the strategies to learn better video feature representation; and 3) the QA pairs of our dataset are bilingual with alignment. This property may inspire multi-lingual VideoQA studies.

Since the models studied here represent the baseline, there remains significant room for improvement. For example, by introducing a more advanced video feature representation model that can learn better discriminative visual features or introducing a more powerful VideoQA model that can perform accurate spatio-temporal reasoning. Furthermore, auxiliary information on ActivityNet, \emph{e.g.}, dense captions \cite{krishna2017dense} can be utilized to help better understanding the fine-grained semantics of videos.

\section*{Acknowledgments}
This work was supported in part by National Natural Science Foundation
of China under Grant 61702143, Grant 61836002, Grant 61622205 and Grant 61602405, and in part by the China Knowledge Centre for Engineering Sciences and Technology, and in part by the Australian Research Council Projects under Grant FL-170100117, Grant DP-180103424 and Grant IH-180100002.

\bibliographystyle{aaai}
%\fontsize{9.5pt}{10.5} \selectfont
\bibliography{aaai2019}

\end{document}